%% file: bmvc.tex
\documentclass{bmvc2k}

\usepackage{tabu}
\usepackage[table]{xcolor}
\usepackage{caption}
\usepackage{setspace} 
\usepackage{pdfpages}

\usepackage{algorithm}
\usepackage{algpseudocode}
\algnewcommand\algorithmicinput{\textbf{Input:}}
\algnewcommand\INPUT{\item[\algorithmicinput]}
\algnewcommand\algorithmicoutput{\textbf{Output:}}
\algnewcommand\OUTPUT{\item[\algorithmicoutput]}
\algnewcommand\STATE{\State}
\algnewcommand\REPEAT{\Repeat}
\algnewcommand\UNTIL{\Until}

\title{Spotlight the Negatives:\\A Generalized Discriminative Latent Model}

\addauthor{Hossein Azizpour}{azizpour@kth.se}{1}
\addauthor{Mostafa Arefiyan}{mostafa@brown.edu}{2}
\addauthor{Sobhan Naderi Parizi}{sobhan@brown.edu}{2}
\addauthor{Stefan Carlsson}{stefanc@csc.kth.se}{1}

\addinstitution{
 CVAP\\
 Royal Institute of Technology (KTH)\\
 Stockholm, Sweden
}
\addinstitution{
 LEMS\\
 Brown University\\
 Providence, USA
}

\runninghead{Azizpour \etal}{Spotlight the Negatives}
\def\ie{\emph{i.e}\bmvaOneDot}
\def\eg{\emph{e.g}\bmvaOneDot}

\def\etal{\emph{et al}\bmvaOneDot}

\DeclareMathOperator*{\argmin}{arg\,min}

\DeclareMathOperator*{\argmax}{arg\,max}

\newcommand{\lv}[3]{
	\def\latentvar{notdefined}
	\ifthenelse{\equal{#1}{+}}
		{\def\latentvar{a}}
		{\def\latentvar{b}}
	\ifthenelse{\equal{#3}{}}
		{\latentvar_{#2}}
		{
			\ifthenelse{\equal{#2}{}}
				{\latentvar^{(#3)}}
				{\latentvar_{#2}^{(#3)}}
		}
}
\newcommand{\mathenv}[1]{{\small
\vspace{-0.0cm}
\begin{align}
#1
\end{align}
\vspace{-0.3cm}
}\\}

\newcommand{\captionfonotsize}6

\begin{document}
\maketitle
\input{Abstract.tex}
\vspace{-0.5cm}
\input{Introduction.tex}
\vspace{-0.5cm}
\input{LVM.tex}
\vspace{-0.5cm}
\input{GLVM.tex}
\input{Training.tex}
\input{Connection.tex}
\input{Demonstrators.tex}

\input{VOCresults.tex}
\input{Experiments.tex}
\vspace{-0.4cm}
\input{Conclusion.tex}
\paragraph{Acknowledgement.}We'd like to thank Pedro Felzenszwalb for the fruitful discussions.
\clearpage
{\small
\bibliography{egbib}
}
\clearpage
\includepdf[pages=-]{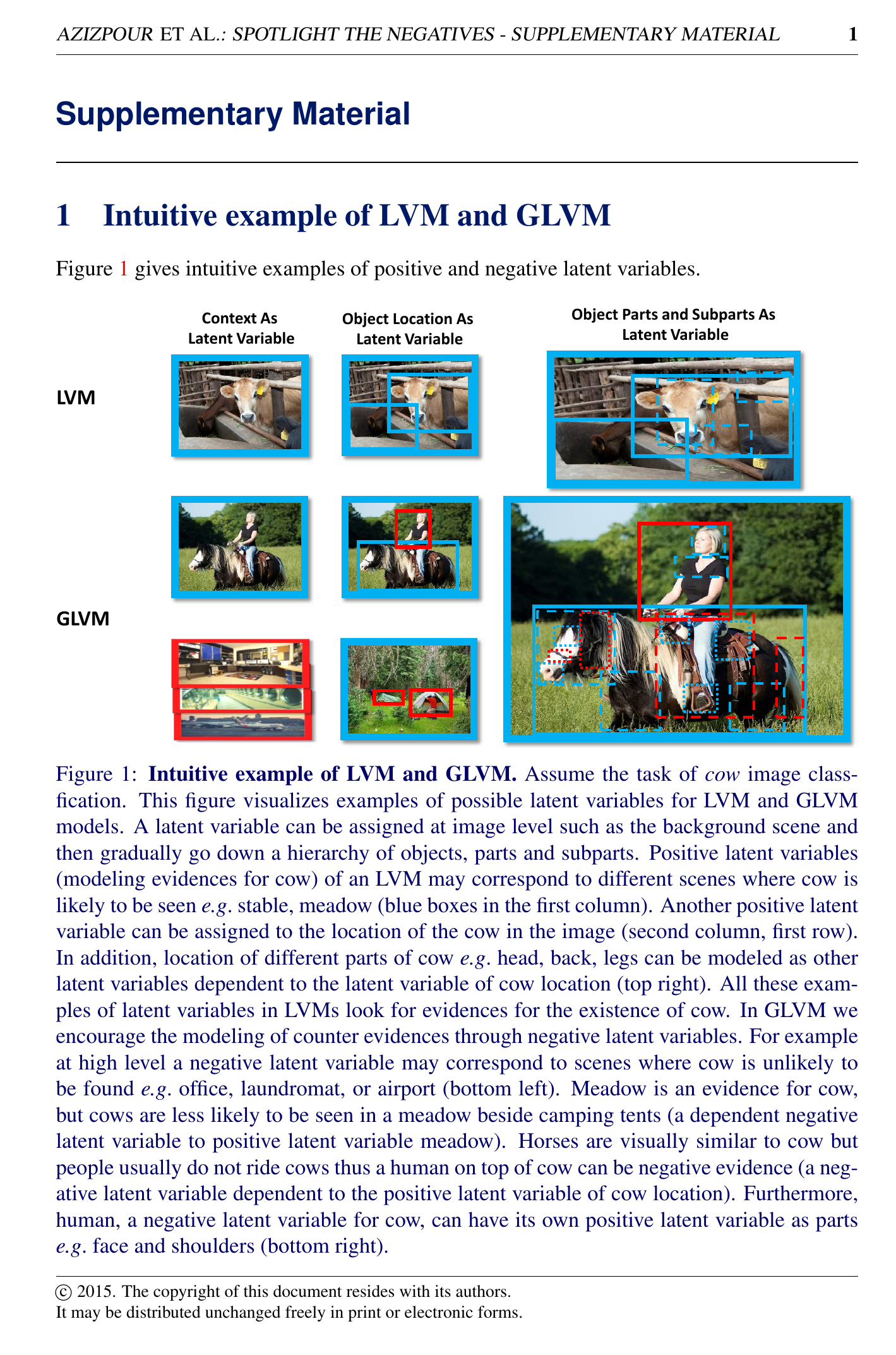}
%
\end{document}

%% file: Abstract.tex
\vspace{-0.6cm}
\begin{abstract}
Discriminative latent variable models (LVM) are frequently applied to various visual recognition tasks. In these systems the latent (hidden) variables provide a formalism for modeling structured variation of visual features. Conventionally, latent variables are defined on the variation of the foreground (positive) class. In this work we augment LVMs to include \textit{negative} latent variables corresponding to the background class. We formalize the scoring function of such a generalized LVM (GLVM). Then we discuss a framework for learning a model based on the GLVM scoring function. We theoretically showcase how some of the current visual recognition methods can benefit from this generalization. Finally, we experiment on a generalized form of Deformable Part Models with negative latent variables and show significant improvements on two different detection tasks.
\end{abstract}

%% file: Introduction.tex
\section{Introduction}
\label{sec:intro}
Discriminative latent variable models (LVM) are frequently and successfully applied to various visual recognition tasks \cite{FelzenszwalbMR08, GirshickIDM15, YuJ09, Yang0VM12, RazaviGKG12, SongWJZ13, PariziVZF15}. In these systems the latent (hidden) variables provide a formalism for modeling structured variation of visual features. These variations can be the result of different possible object locations, deformations, viewpoint, subcategories, etc. Conventionally, these have all been defined only based on the foreground (positive) class. We call such latent variables ``positive''. In this work we introduce generalized discriminative LVM (GLVM) which use both ``positive'' and ``negative'' latent variables. Negative latent variables are defined on the background (negative) class and provide \emph{counter evidence} for presence of the foreground class. They can, for instance, learn mutual exclusion constraints, model scene subcategories where the positive object class is unlikely to be found, or capture specific parts which potentially indicate the presence of an object of a similar but not the same class. \\

\noindent The objective of the proposed framework is to learn a model which maximizes the evidence (characterized by positive latent variables) for the foreground class and at the same time minimizes the counter evidence (characterized by negative latent variables) lying in background class. Thereby GLVM empowers the latent variable modeling by highlighting the role of negative data in its own right. An interesting analogy is with game theory; when playing a game a simple strategy is to maximize your score. This resembles LVM's objective which is to maximize evidence for the presence of the foreground class. A better strategy, however, is to maximize your scores at the same time as minimizing the opponents' scores. This is analogous to GLVM's objective which takes into account counter evidence emerging from negative latent variables while looking for evidence from the foreground class. The score of the opponents is counter evidence in our hypothetical example. \\

\noindent  The concept of \emph{negative parts} was noted in~\cite{PariziVZF15}. However, \cite{PariziVZF15} focuses on automatic discovery and optimization of a part based model with negative parts. In this paper we extend the notion of negative parts to negative latent variables and propose a framework for defining models that use both positive and negative latent variables. \\

\noindent Many different modeling techniques benefit from latent variables. Felzenszwalb \etal~\cite{FelzenszwalbMR08} introduced latent SVM to train deformable part models, a state of the art object detector~\cite{GirshickIDM15}. Yu and Joachims~\cite{YuJ09} transferred the idea to structural SVMs. Yang \etal~\cite{Yang0VM12} proposed a kenelized version of latent SVM and latent structural SVM and applied it to various visual recognition tasks. Razavi \etal~\cite{RazaviGKG12} introduced latent variables to the Hough transform and achieved significant improvements over a Hough transform object detection baseline. And-Or trees use latent variable modeling for object recognition and occlusion reasoning~\cite{SongWJZ13}. \\

\noindent In this paper we introduce a novel family of models which generalizes discriminative latent variable models.
We review LVMs in Section~\ref{sec:LVM}, formulate our generalization of LVMs in Section~\ref{sec:GLVM}. Then, we discuss training of GLVMs and propose an algorithm for learning the parameters of a GLVM in Section~\ref{sec:training}. In Section~\ref{sec:demonstrators} we discuss how various computer vision models can benefit from GLVM. Finally, in Section~\ref{sec:experiments} we experiment on generalized DPMs.

%% file: LVM.tex
\vspace{-0.25cm}
\section{Discriminative Latent Variable Models (LVM)}
\label{sec:LVM}
\vspace{-0.2cm}
\indent Consider a supervised classification problem provided with a set of $n$ training examples $X\in \mathcal{R}^{d\times n}$ and their corresponding labels $Y\in \mathcal{Y}^{n}$. The goal of discriminative learning is then to learn a scoring function $S_w(x,y)$,  parametrized by $w$, that ranks a data point $x_i$ and its true label $y_i$ higher than the rest of labels. For a linear scoring function this is written as follows:
\mathenv{
S_w(x_i,y_i) > S_w(x_i, y) \qquad \forall y\neq y_i \\
S_w(x_i,y) = w^{T}\phi(x_i, y) \label{eq:score_supervised}
}
Here $\phi(x, y)$ is the representation function (e.g. Histogram of Oriented Gradients, ConvNet representation, etc.). Since the scoring function of (\ref{eq:score_supervised}) is linear in $w$ any invariance to nonlinear object class deformation should be reflected in the representation function $\phi$. This requires a very rich feature encoding. LVMs model known structured variations of the object class by using various latent variables $\lv{+}{}{}$ corresponding to different sources of variation. Each latent variable models a source of variation and accordingly alters the representation $\phi$ so that it best matches the model parameters $w$. Thus, all latent variables in an LVM are maximized over. For instance, in weakly supervised object detection location of the object in an image is modeled by a latent variable. The scoring function of an LVM takes this form:
\vspace{-0.38cm}
\mathenv{
S_w(x_i,y) = \max_{\lv{+}{}{}}w^{T}\phi(x_i, y, \lv{+}{}{})
\label{eq:LVM_score}
}\\
\noindent In some cases the set of labels $\mathcal{Y}$ comprises of only known classes. In open-set classification, however, one of the possible labels in $\mathcal{Y}$ refers to an \emph{unknown class}. The simplest form of open-set classification is binary classification where samples are classified into the class of interest where $y=+1$ (\eg cat) or anything else where $y=-1$ (\eg non-cat) which are so called \textit{foreground} and \textit{background} respectively. In this work we are interested in open-set classification problems which suit many real-world applications better. In the next section we show how the latent variables in the scoring function of (\ref{eq:LVM_score}) can be altered to ``model'' the negative data in its own right and thereby enrich the expressive power of discriminative latent variable models. In the rest of the paper we focus on the binary classification problem.\\

%% file: GLVM.tex
\vspace{-0.3cm}
\section{Generalized LVM with Negative Modeling (GLVM)}
\label{sec:GLVM}
\vspace{-0.1cm}
 The scoring function in discriminative LVM is usually constructed in such a way that it looks for ``evidence'' for the foreground class, and the weakness (or lack) of such evidence indicates that the input is from the background class. For instance, in detecting \emph{office} scenes, the location of a work desk (evidence for office) can be modeled by a latent variable. \\
\noindent We generalize LVMs by equipping them with latent variables that look for ``counter evidence'' for the foreground class. For example, detecting a stove (evidence for kitchen) or sofa (evidence for living room) provides counter evidence for office. Such latent variables can also be contextual, for example for a \textit{cow} detector, presence of saddle/rider/aeroplane counts as counter evidence whereas presence of meadow/stable/milking parlor counts as evidence. For visual examples refer to Section 1 in supplementary material.\\
\noindent We call the latent variables that collect evidence for the foreground class \textit{positive latent variables} and the latent variables that collect counter evidence for the foreground class \textit{negative latent variables} and denote them by $\lv{+}{}{}$ and $\lv{-}{}{}$ respectively throughout the paper.\\

\noindent \textbf{Representation function.}  Before we go further, let's explain how our representation function $\phi$ is constructed. We assume the model parameters (associated with \textit{all} latent variables) are concatenated into a single vector $w$. Moreover, $w$ and its corresponding feature vector $\phi(x_i,\lv{+}{1}{},\lv{-}{1}{},\lv{+}{2}{},\lv{-}{2}{},...)$ have the same dimensionality. Every latent variable $\lv{+}{k}{}$ or $\lv{-}{k}{}$ encodes its own place in $\phi()$. The places for different latent variables do not overlap. When some latent variables are omitted from the inputs of $\phi()$, their place is filled with zero vectors. These assumption are made for the ease of notation in the upcoming formulations. For clarity we take three steps towards defining the generic form of GLVM scoring function.\\
\input{Figure_Train.tex}
\noindent\textbf{Simple Form.} The simplest form of a GLVM scoring function involving two independent sets of positive and negative latent variables is ($\mathcal{Z}$ is the set of possible latent assignments):
\mathenv{
S_w(x_i, y) = \max_{\lv{+}{}{} \in \mathcal{Z}^{+}}w^{T}\phi(x_i, y, \lv{+}{}{}) \;\; - \;\; \max_{\lv{-}{}{} \in \mathcal{Z}^{-}}w^{T}\phi(x_i, y, \lv{-}{}{})
\label{eq:GLVM_score}
}
Note that the second $\max$ cannot be absorbed in the first $\max$. Thus, the GLVM scoring function is more general than that of LVM. In particular, (\ref{eq:GLVM_score}) reduces to (\ref{eq:LVM_score}) when ${|\mathcal{Z}^{-}| \leq 1}$.
One can characterize positive and negative latent variables according to the way they impact the value of the scoring function. Each assignment of a latent variable has a corresponding value associated to it. For instance, in the deformable part models (DPMs) latent variables indicate the location of the object parts~\cite{FelzenszwalbMR08}. Each assignment of a latent variable in a DPM corresponds to a placement of a part which in turn is associated with a value in the score map of the part. A positive (negative) latent variable is one whose associated value is positively (negatively) correlated with the final value of the scoring function. This should not be confused with negative entries in the model vector $w$. In a linear model with the score function $S(x) = w^{T} \phi(x)$ the value of the $d$-th dimension of the feature vector (\ie $\phi(x)_d$) is negatively correlated with $S(x)$ when $w_d < 0$. Yet, negative latent variables are fundamentally different from this. In particular, a negative latent variable cannot be obtained by flipping the sign in some dimensions of the parameter vector and using a positive latent variable instead. The value associated with a negative latent variable (\ie $max_{\lv{-}{}{}} w^{T} \phi(x, \lv{-}{}{})$) is a convex function of $w$ whereas the value associated to negative entries in $w$ (\ie $w_d \phi(x)_d$) is linear in $w$.\\

\noindent\textbf{Dependent Form.} The value of the GLVM score can be thought of as the value of the best strategy in a 2-player game. To show this we need to modify the notation as follows:
\vspace{-0.1cm}
\mathenv{
\nonumber S_w(x_i) &= \max_{\lv{+}{}{} \in \mathcal{Z}^{+}}w^{T}\phi(x_i, \lv{+}{}{}) +  \min_{\lv{-}{}{} \in \mathcal{Z}^{-}}-w^{T}\phi(x_i, \lv{-}{}{}) \\
\nonumber &= \max_{\lv{+}{}{} \in \mathcal{Z}^{+}}\min_{\lv{-}{}{} \in \mathcal{Z}^{-}} w^{T}\phi(x_i, \lv{+}{}{}) - w^{T}\phi(x_i, \lv{-}{}{})\\
&= \max_{\lv{+}{}{} \in \mathcal{Z}^{+}}\min_{\lv{-}{}{} \in \mathcal{Z}^{-}} ~ w^{T}\phi(x_i, \lv{+}{}{}, \lv{-}{}{})
\label{eq:GLVM_score2}
}
In the first step we dropped $y$ from the notation since we will be focusing on binary classification.
In the second step $\min$ is pulled back. For the last step we let $\phi(x_i, \lv{+}{}{}, \lv{-}{}{}) = \phi(x_i, \lv{+}{}{}) - \phi(x_i, \lv{-}{}{})$ although any feature function that \textit{depends on both} $\lv{-}{}{}$ and $\lv{+}{}{}$ would do. So, (\ref{eq:GLVM_score2}) is more general than (\ref{eq:GLVM_score}). Note that the game theory analogy of minimax is now more apparent in the new formulation of (\ref{eq:GLVM_score2}). \\

\noindent\textbf{Canonical Form.} The scoring function in (\ref{eq:GLVM_score2}) is linear when the latent variables $\lv{+}{}{}$ and $\lv{-}{}{}$ are fixed. One can imagine augmenting this linear function with more positive and negative latent variables; the same way that we extended (\ref{eq:score_supervised}) to get to (\ref{eq:GLVM_score2}). This creates multiple \textit{interleaved} negative and positive latent variables. This way, we can make a hierarchy of alternating latent variables recursively. This general formulation has interesting ties to compositional models, like object grammar~\cite{FelzenszwalbM10}, and high-level scene understanding. For example, we can model scenes according to the presence of some objects (positive latent variable) and the absence of some other objects (negative latent variable). Each of those objects themselves can be modeled according to presence of some parts (positive) and absence of others (negative). We can continue the recursion even further. The GLVM scoring function, after being expanded recursively, can be written in the following general form:
\vspace{-0.1cm}
\mathenv{
S_w(x_i) = \max_{\lv{+}{1}{}}\min_{\lv{-}{1}{}} ~ \max_{\lv{+}{2}{}}\min_{\lv{-}{2}{}} ~ ... ~ \max_{\lv{+}{K}{}}\min_{\lv{-}{K}{}} 
~ w^{T} \phi \left( x_i, (\lv{+}{k}{}, \lv{-}{k}{})^{K}_{k=1} \right)
\label{eq:GLVM_score3}
}
We call this the \emph{canonical form} of a GLVM scoring function. Note that any other scoring function that is a linear combination of $\max$'s and $\min$'s of linear functions can be converted to this canonical form by simple algebraic operations. Moreover, (\ref{eq:GLVM_score3}) with $K>1$ and non empty $\lv{+}{k}{}$ and $\lv{-}{k}{}$ cannot be simplified into (\ref{eq:GLVM_score2}), in general. Its proof follows the \textit{max-min inequality}~\cite{Boyd2004}. See Figure \ref{fig:tree} for an example.

%% file: Figure_Train.tex
    \begin{figure}
    \centering
    		\includegraphics[trim = 0mm 100mm 0mm 23mm, clip, width=115mm]{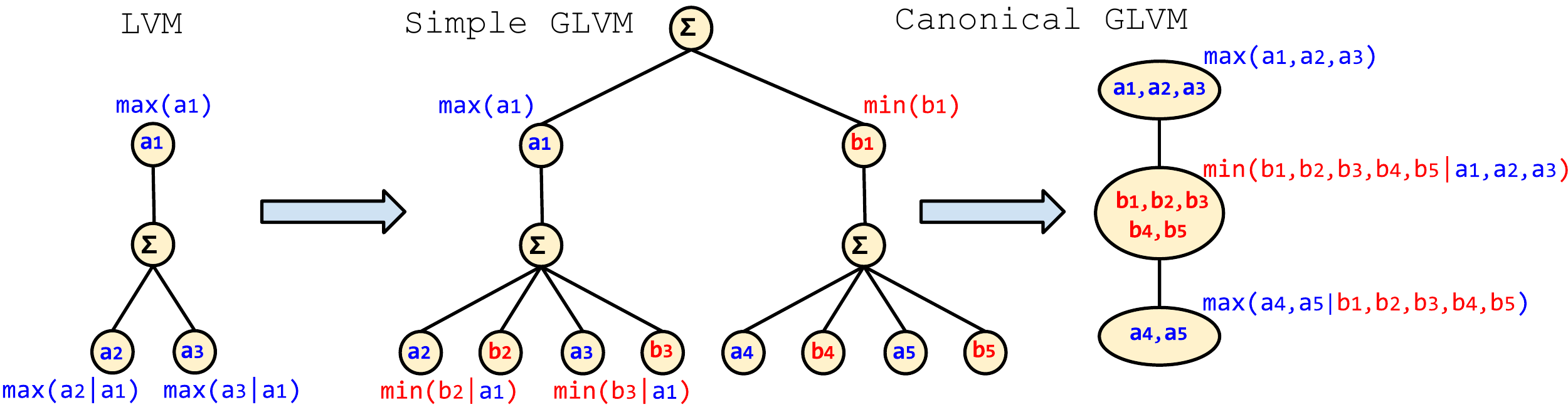}
            \caption{{\fontsize{\captionfonotsize pt}{0pt}\selectfont\color{bmv@captioncolor} Example of an LVM scoring function on three latent variables (left), a Simple GLVM scoring function (middle), and a GLVM in its canonical form. The models become more and more complex from left to right.}}
            \label{fig:tree}
            \vspace{-0.5cm}
	\end{figure}\\

%% file: Training.tex
\section{Training GLVMs for Classification}
\label{sec:training}
In this section we propose an algorithm for training GLVMs to do binary classification. We focus on SVM training objective and use hinge loss. That being said, the same idea can be used to optimize any training objective that involves a linear combination of convex and/or concave functions of GLVM scores; \eg SVM with squared hinge loss. For the ease of notation we drop the subscript $w$ in $S_w$ in the rest of this section. \\

\noindent Let $D=\{ (x_i, y_i) \}_{i=1}^n$  denote a set of $n$ labeled training examples. Each training example is a pair $(x_i, y_i)$ where $x_i$ is the input data and $y_i \in \{+1, -1\}$ is the ground-truth label associated with it. SVM training objective can be written as follows:
\mathenv{
	O(w) = \frac{1}{2} ||w||^2 + C \sum_{i=1}^n \max \left( 0, 1 - y_i S(x_i) \right) \label{eq:svm_objective}
}
where $S(x_i)$ is a scoring function. Training a classifier requires finding the model $w^*$ that minimizes the function \ie $w^* = \argmin_{w} O(w)$. In most practical applications, however, the objective function $O$ is complex (in particular, non-convex). This makes it hard to find the global minimizer. Thus, we usually have to resort to local minima of the objective function. \\

\noindent The concave-convex procedure (CCCP)~\cite{YuilleR03} is a general framework for optimizing non-convex functions. For minimization, in each iteration the concave part is replaced by a hyperplane tangent to it at the current solution. The result is a convex function. Local minima of a convex function are global minima. Thus, one can use simple methods such as gradient descent to minimize a convex function. \\

\noindent We train GLVMs using an iterative algorithm that is very similar to CCCP~\cite{YuilleR03}. But, unlike CCCP, we do not linearize the concave part of the objective function. Our training algorithm alternates between two steps: 1) constructing a convex upper bound to the training objective $O$ and 2) minimizing the bound. Let $B_t(w)$ denote the bound constructed at iteration $t$. Also let $w_t = \argmin_w B_t(w)$ be the minimizer of the bound. All bounds must touch the objective function at the solution of the previous bound; that is $B_t(w_{t-1}) = O(w_{t-1}), \forall t$. This process is guaranteed to converge to a local minimum or a saddle point of the training objective $O$. \\

\noindent To construct bounds $B_t$, as we shall see next, we only need to construct a concave lower bound $\hat{S}(x_i)$ and a convex upper bound $\check{S}(x_i)$  to the scoring function of (\ref{eq:GLVM_score3}). We define $B_t$ as follows:
\mathenv{
	B_t(w) = \frac{1}{2} ||w||^2 + C \sum_{\substack{i=1 \\ y_i = -1} }^n \max \left( 0, 1 + \check{S}_t(x_i) \right) + C \sum_{\substack{i=1 \\ y_i = +1} }^n \max \left( 0, 1 - \hat{S}_t(x_i) \right)
 \label{eq:svm_bound}
}
In the next section we will explain how $\check{S}(x_i)$ and $\hat{S}(x_i)$ are constructed for GLVMs. Algorithm 1 in the supplementary material summarizes all steps in our training procedure.
\\

\noindent\textbf{Convex Upper Bound and Concave Lower Bound.}
Here we explain how we obtain $\check{S}_t(x_i)$ and $\hat{S}_t(x_i)$ when GLVM scoring function (\ref{eq:GLVM_score3}) is used. We fix all the negative latent variables in $S(x_i)$ to get $\check{S}_t(x_i)$ and fix all the positive latent variables to get $\hat{S}_t(x_i)$. \\

\noindent Let $w_t$ denote the model at iteration $t$ and let $\lv{+}{}{}=(\lv{+}{1}{}, \dots, \lv{+}{K}{})$ be an arbitrary assignment of the positive latent variables. Given $\lv{+}{}{}$, we denote the fixed values for the negative latent variables on image $x_i$ by $\lv{-}{}{i} (\lv{+}{}{}) = (\lv{-}{1}{i}, \dots, \lv{-}{K}{i})$ and define it as in (\ref{eq:fixed_negative_parts}). Similarly, we define $\lv{+}{}{i} (\lv{-}{}{}) = (\lv{+}{1}{i}, \dots, \lv{+}{K}{i})$ as in (\ref{eq:fixed_positive_parts}).
\mathenv{
	\lv{-}{}{i}(\lv{+}{}{}) = \argmin_{\lv{-}{1}{}, \dots, \lv{-}{K}{}} ~ w_t^{T} \phi \left( x_i, (\lv{+}{k}{}, \lv{-}{k}{})^{K}_{k=1} \right) \label{eq:fixed_negative_parts} \\
	\lv{+}{}{i}(\lv{-}{}{}) = \argmax 	_{\lv{+}{1}{}, \dots, \lv{+}{K}{}} ~ w_t^{T} \phi \left( x_i, (\lv{+}{k}{}, \lv{-}{k}{})^{K}_{k=1} \right) \label{eq:fixed_positive_parts}
}
Note that (\ref{eq:fixed_negative_parts}) uses a $\min$ and (\ref{eq:fixed_positive_parts}) uses a $\max$. Also, note that in both equations the fixed assignments are computed according to $w_t$.  Finally, we define $\check{S}_t(x_i)$ and $\hat{S}_t(x_i)$ as follows:
\mathenv{
	\check{S}_t(x_i) &= \max_{\lv{+}{1}{}} ~ \max_{\lv{+}{2}{}} ~ \dots ~ \max_{\lv{+}{K}{}}~ w^{T} \phi \left( x_i,  \left(\lv{+}{k}{}, \lv{-}{k}{i}(\lv{+}{}{}) \right)^{K}_{k=1} \right) \label{eq:GLVM_score_convexUB} \\
	\hat{S}_t(x_i) &= \min_{\lv{-}{1}{}} ~ \min_{\lv{-}{2}{}} ~ \dots ~ \min_{\lv{-}{K}{}}~ w^{T} \phi \left( x_i,  \left( \lv{+}{k}{i}(\lv{-}{}{}), \lv{-}{k}{} \right )^{K}_{k=1} \right) \label{eq:GLVM_score_concaveLB}
}
(\ref{eq:GLVM_score_convexUB}) is convex because it is a $\max$ of linear functions and it is an upper bound to $S(x)$ because the $\min$ functions are replaced with a fixed value. Similarly for (\ref{eq:GLVM_score_concaveLB}).\\
For discussions regarding the dependency structure of the latent variables and how it affects the scoring and inference functions refer to Section 2
in the supplementary material.

%% file: Connection.tex
\section{Connection to existing Models}
\label{sec:connection}
\textbf{LSSVM: }An interesting connection is to latent structural SVMs (LSSVM) \cite{YuJ09} where the scoring function is given by,
\mathenv{
S(x_i, y_i) = \max_{h} w^{T}\phi(x_i,y_i,h)
}
In this form, the formulation of scoring function does not make use of negative latent variables as defined in GLVM. In LSSVM objective function the individual loss induced by a sample $(x_i,y_i)$ is calculated as below,
\mathenv{
L(x_i,y_i) = \max_{(y,h)}[w^{T}\phi(x_i,y,h)+\Delta(y_i,y,h)]-\max_h w^{T}\phi(x_i,y_i,h)
}
Note that the representation function takes as input both class $y$ and latent variable $h$. Now, assume a binary classification ($y\in\{-1,1\}$) framework with a symmetric loss function $\Delta$ independent of $h$.
Then, the scoring and loss function can be reformulated as,
\mathenv{
&S'(x_i) = \max_{h}w^{T}\phi(x_i,+1,h)-\max_{h} w^{T}\phi(x_i,-1,h) \label{eq:LSSVM_newscore0} \\
&L(x_i,y_i) = \max(0,\Delta(+1,-1)- y_i S'(x_i)) \label{eq:LSSVM_newscore}
}
Note that (\ref{eq:LSSVM_newscore0}) is similar to the shallow structure of GLVMs in (\ref{eq:GLVM_score}). 
This shows how negative latent variables can potentially emerge from LSSVMs.\\

\noindent\textbf{Mid-level features based recognition: }Many recent scene image classification techniques pre-train multiple mid level features (parts) \cite{JunejaVJZ13, DoerschGE13, EndresSJH13} to represent an image. They construct the representation of an image by concatenating the maximum response of each part in the image and then train a linear classifier on top of those representations. Naderi \etal \cite{PariziVZF15} have shown that training a classifier on those representations effectively corresponds to the definition of \textit{negative parts}. In these works, the score of a part at an image is \textit{maximized} over different locations and thus its position and scale are, in essence, treated as a latent variables. In that respect, a negative weight in the linear classifier for such a latent variable can be translated to a \textit{negative latent variable} (\ie its maximum score contributes negatively to the score of an image). 
The training procedure for the scene classification models of \cite{PariziVZF15,JunejaVJZ13, DoerschGE13, EndresSJH13} is similar to the shallow version of the GLVM framework proposed in this paper.\\

\noindent\textbf{ConvNet: }\cite{GirshickIDM15} argues that the kernels in Convolutional Networks (ConvNet) can be interpreted as parts in part  based models and shows that DPM is effectively a ConvNet. However, since ConvNets have the extra ability of learning weights  (potentially negative) over these multiple kernels it is modeling counter evidences in a similar form as GLVM.

%% file: Demonstrators.tex
\section{Showcases}
\label{sec:demonstrators}
To elicit the implications of the proposed generalization of LVMs, in this section, we review a few different state of the art computer vision models and demonstrate how they can benefit from the proposed framework. We consider adding negative latent variables to deformable part models, And-Or trees, and latent Hough transform and discuss the implications of it.\\
\input{DPM.tex}
\input{AndOr.tex}
\input{Hough_short.tex}

%% file: DPM.tex
\\

\noindent\textbf{Show Case 1: Deformable Part Models.}
\label{sec:dpm}
Deformable Part Model (DPM) \cite{FelzenszwalbMR08} is an object detector that models an object as a collection of deformable parts. DPM uses multiple mixture components to model viewpoint variations of the object. The location of the parts (denoted by $l_j$) and the choice of the mixture component (indexed by $c$) are unknown and thus treated as latent variables.  The scoring function of a DPM with $n$ parts can be written as follows:
\mathenv{
S(x) = \max_{c=1}^{k}\sum_{j=1}^{n}\max_{l_j}w_{c,j}^{T}\phi(x,l_j,c)
\label{DPM_score}
}
We extend DPMs to include negative parts. We refer to this as generalized DPM (GDPM). For instance, when detecting cow, positive parts in DPM include head, back, legs of a cow. In GDPM one can imagine several candidates for negative parts. For example, since horses appear frequently in the high scoring false positives of cow detectors, saddle or horse-head may be good negative parts for a cow detector. We define  GDPM scoring function as:
\mathenv{
S(x) = \max_{c=1}^{k}\Big[\sum_{j^{+}=1}^{n}\max_{l_{j^+}}w^{T}\phi(x,l_{j^+},c)
-\sum_{j^{-}=1}^{m}\max_{l_{j^-}}w^{T}\phi(x,l_{j^-},c)\Big]
\label{DPM_score}
}
Following the recursion of GLVM, the outer max can also be cloned with a negative counterpart. In that respect, the negative \textit{components} would look, for instance, for different viewpoints of a horse with its own positive and negative parts. We implemented GDPM and evaluated it on two different datasets. We discuss the experimental results in Section \ref{sec:experiments}.

%% file: AndOr.tex
\\

\noindent\textbf{Show Case 2: And-Or trees.}
\label{sec:andor}
And-Or trees \cite{SongWJZ13} are hierarchical models that have been applied to various visual recognition tasks. And-Or trees are represented by a tree-structured graph. Edges in the graph represent dependencies and nodes represent three different operations. \textit{Terminal} nodes ($T_i$) are directly applied to the input image. A typical example of a terminal node is a template applied at a specific location of the image. \textit{Or} nodes ($O_i$) take the maximum value among their children. An \textit{Or} node can model, for example, the placement of terminal nodes or the choice of mixture components. \textit{And} nodes ($A_i$) aggregate information by summing over their children. 
The scoring functions of these nodes are defined as follows:
\mathenv{
S(T_i|I;w_i) = w_{i}^{T}I(T_i),  \qquad S(A_i|C) = \sum_{c\in C(A_i)} S(c), \qquad S(O_i|C) = \max_{c\in C(O_i)} S(c)
}
$C$ encodes the hierarchy of the nodes and $C(N)$ denotes children of node $N$. In order to compute the score of an input image the scores are propagated from the terminal nodes to the root of hierarchy using dynamic programming. Or nodes find the most plausible interpretation, And nodes aggregate information, and terminal nodes collect positive evidence for the foreground class.
Generalizing an And-Or tree with negative latent variables introduces a new type of node which we call a \emph{Nor} node ($N_i)$. Nor nodes collect counter evidence for the class of interest. We define the scoring function of a Nor node as follows:
\mathenv{
S(N_i|C) = \min_{c\in C(N_i)} S(c)
}

%% file: Hough_short.tex
\noindent\textbf{Show Case 3: Latent Hough  Transform.}
\label{sec:hough}
Hough transform (HT) has been used as a non-parametric voting scheme to localize objects. In Implicit Shape Models \cite{LeibeLS08} the score of a hypothesis $h$ (\eg object location and scale) in an image $x$ is calculated by aggregating its compliance to each \textit{positive} training images. Let $D_p$ denote the set of positive training images. Also let $S(x, h | x_i)$ denote the vote of a positive training image $x_i \in D_p$ in favor of hypothesis $h$. The scoring function of HT is defined as follows:
\mathenv{
&S_{HT}(x,h) = \sum_{i\in D_p} S(x,h|x_i)
}
One of the main issues with HT is the aggregation of votes from inconsistent images. 
For instance, training images of side-view cars contribute to the voting in a test image of a frontal-view car. Razavi \etal \cite{RazaviGKG12} alleviate this issue by introducing latent Hough transform (LHT). LHT groups consistent images (\eg exhibiting same viewpoints) together using a latent variable $z \in Z$. Let $\phi(x,h)$ denote the vector of votes for hypothesis $h$ obtained from all positive training images \ie $\phi(x, h)_i = S(x, h | x_i)$. LHT assigns weight $w_{z, i}$ to the positive training image $x_i \in D_p$ indicating its relevance to the selected mixture component $z$. Scoring function of an LHT parameterized by $w$ is defined as follows:
\mathenv{
&S_{LHT}(x,h) = \max_{z \in Z} w_z^{T}\phi(x,h)
}
In order to reformulate the problem as a GLVM we need to introduce \textit{negative votes}~\cite{Brown83} cast by negative training images in the scoring function of HT. 
\mathenv{
&S_{HT}(x,h) = \sum_{i\in D_p} S(h|x_i)-\sum_{i\in D_n} S(h|x_i)
\label{NHT_score}
}
Then the scoring function of a GLVM generalization of LHT is given by:
\mathenv{
&S_{LNHT}(x,h) = \max_{z^+} w_{z^+}^{T}\phi(x,h) - \max_{z^-} w_{z^-}^{T}\phi(x,h)
}
 It should be emphasized that negative latent variables might be crucial when using negative votes in HT due the high multimodality of negative data distribution. That is negative images are probably more susceptible to the aggregation of inconsistent votes. This might explain the reason why negative voting is not popular for HT based object recognition. Note that the GLVM version of LHT can make use of the negative data in a more effective way.

%% file: VOCresults.tex


\begin{figure}
	\begin{subfigure}{.33\textwidth}
	\vspace{-0.4cm}
	\input{table2.tex}
	\caption{{\fontsize{\captionfonotsize pt}{0pt}\selectfont\color{bmv@captioncolor} AP for animal classes of VOC 2007, comparing DPM with GDPM with different number of parts. Sub (super) indices indicate the number of positive (negative) parts.}}
	\label{tab:voc}
	\end{subfigure}
	\,
	\begin{subfigure}{.64\textwidth}
	\centering
	  		\includegraphics[trim = 9mm 130mm 77mm 9mm, clip,width=0.28\linewidth]{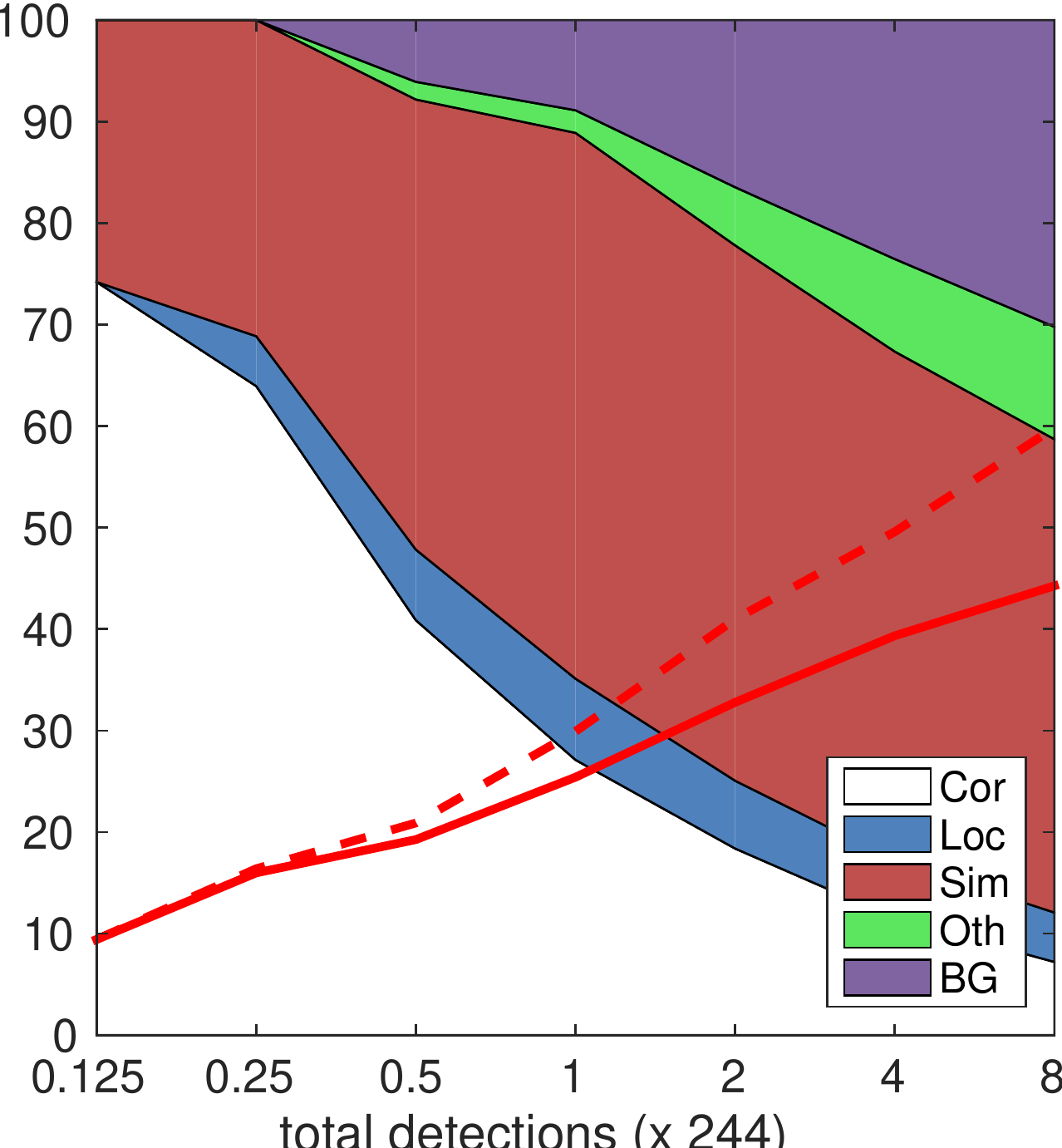}
	  		\includegraphics[trim = 9mm 130mm 77mm 9mm, clip,width=0.28\linewidth]{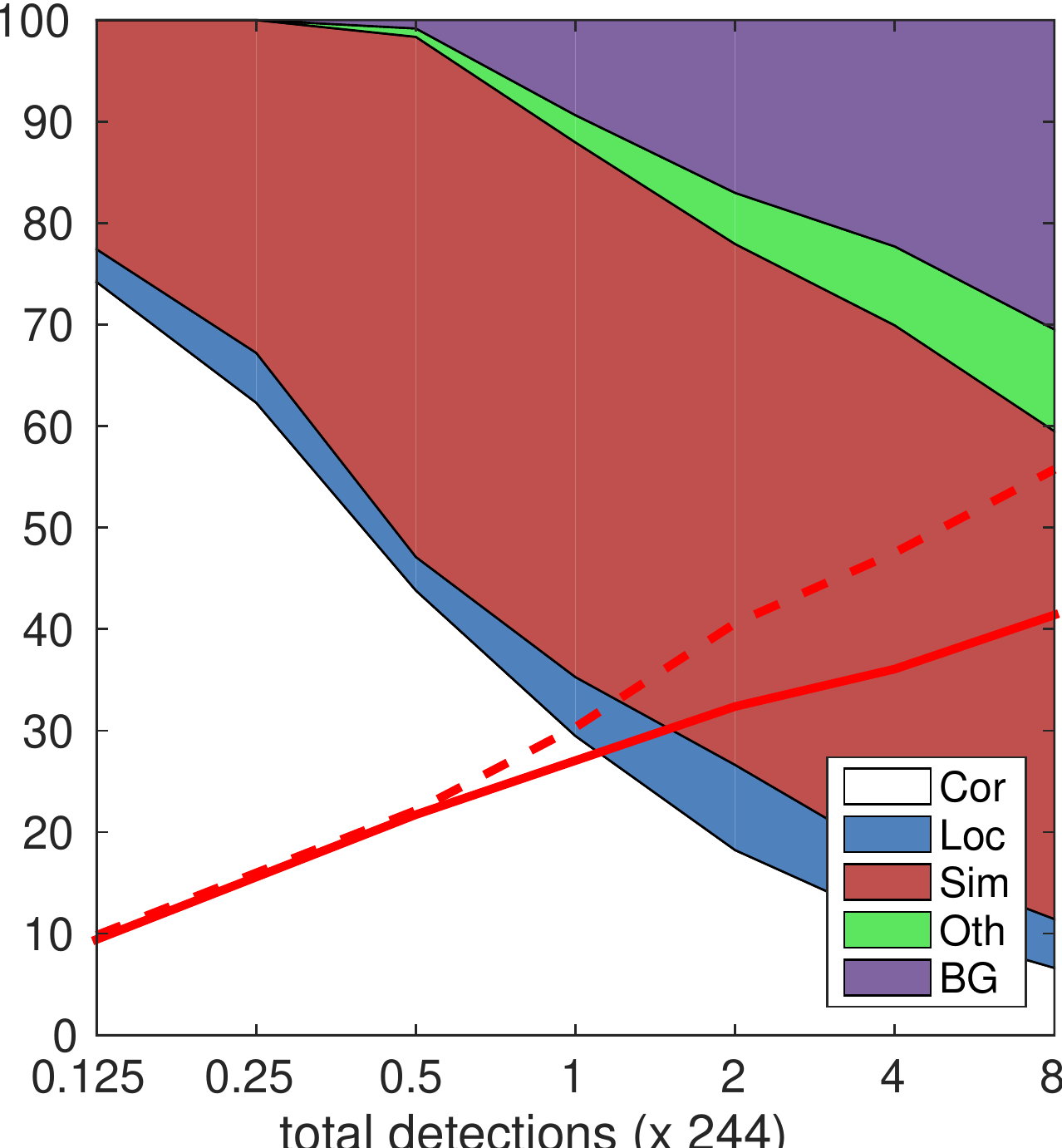}
	  		\includegraphics[trim = 9mm 130mm 77mm 9mm, clip,width=0.28\linewidth]{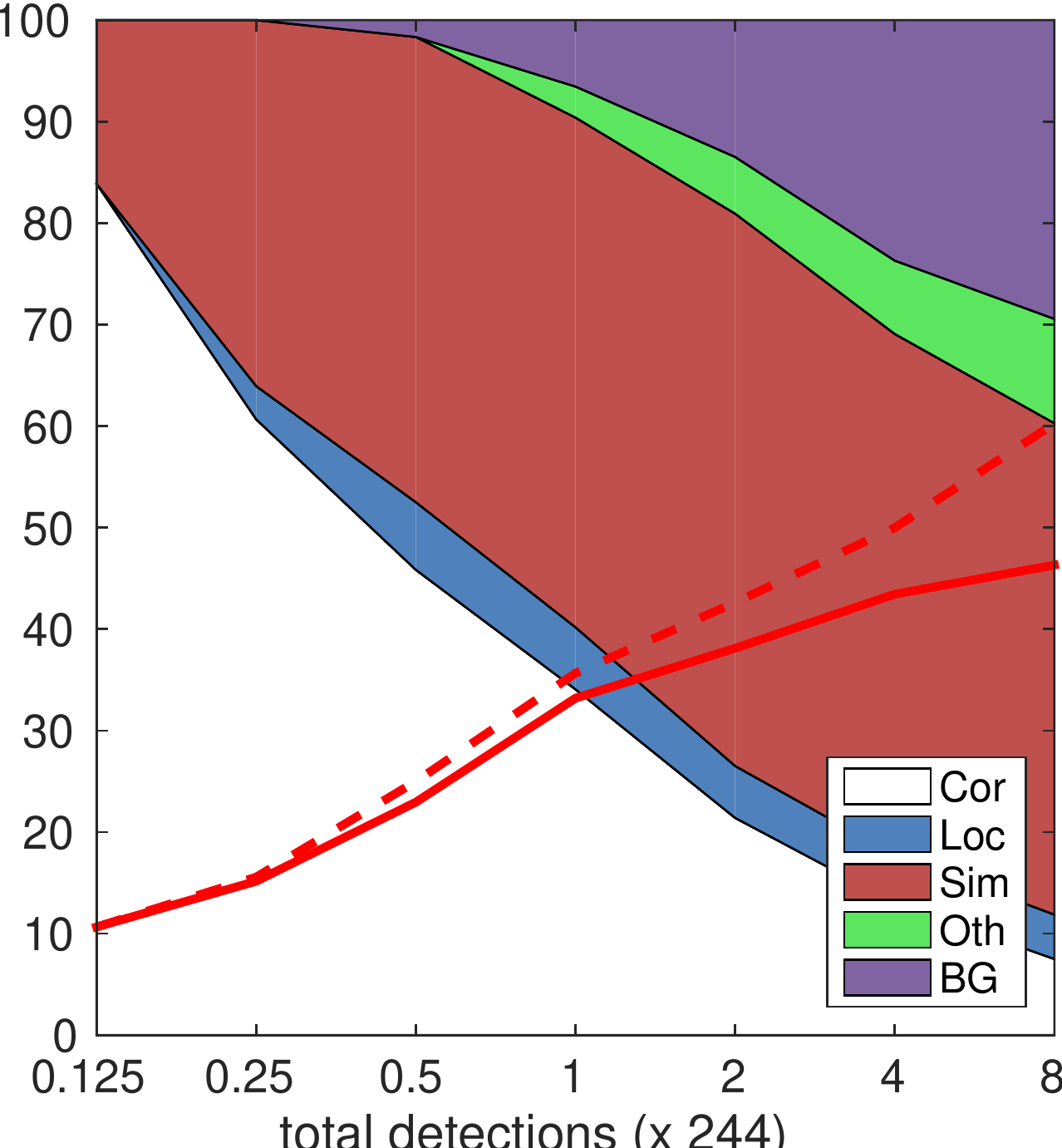}
	   		\vspace{-0.2cm}
	        \caption{{\fontsize{\captionfonotsize pt}{0pt}\selectfont\color{bmv@captioncolor}Analysis of false positives of \textit{cow} using \cite{HoiemCD12}. $DPM$ (left), $GDPM^{2}_6$ (middle) and $GDPM^{1}_{7}$ (right) with similar part initialization.  The volumes show the commulative number of top detections within 5 different categories: true detection (white), localization FP (blue), background FP (purple), similar objects (red), other objects (green). Red (dashed red) line is the percentage of recall at each number of detections using 50\% (10\%) overlap threshold. GDPM with careful initialization (right figure) slightly reduces confusion with similar objects.}}
	        \label{fig:diagnose}
		\end{subfigure}\\
	\caption{{\footnotesize PASCAL VOC 2007}}
	\label{fig:figure-and-table}
	\vspace{-0.45cm}
\end{figure}

%% file: table2.tex
\setlength{\tabcolsep}{1pt}
\begin{tabular}[b]{c | ccc ccc} 
 & \scriptsize{bird} & \scriptsize{cat} & \scriptsize{cow} & \scriptsize{dog} & \scriptsize{horse} & \scriptsize{sheep} \\ \hline
\scriptsize{$\mathbf{DPM_{8}}$} & \scriptsize{\textbf{10.9}} & \scriptsize{16.6} & \scriptsize{21.9} & \scriptsize{\textbf{12.5}} & \scriptsize{56.0} & \scriptsize{18.0}\\
\scriptsize{$\mathbf{GDPM_{7}^{1}}$} & \scriptsize{\textbf{10.9}} & \scriptsize{17.0} & \scriptsize{22.0} & \scriptsize{\textbf{12.5}} & \scriptsize{\textbf{57.2}} & \scriptsize{19.2}\\
\scriptsize{$\mathbf{GDPM_{6}^{2}}$} & \scriptsize{9.5} & \scriptsize{\textbf{18.6}} & \scriptsize{\textbf{23.2}} & \scriptsize{12.2} & \scriptsize{56.0} & \scriptsize{\textbf{19.8}}\\
\scriptsize{$\mathbf{GDPM_{5}^{3}}$} & \scriptsize{10.5} & \scriptsize{13.9} & \scriptsize{21.7} & \scriptsize{11.9} & \scriptsize{54.8} & \scriptsize{19.1}
\end{tabular}

%% file: Experiments.tex
\section{Experiments}
\label{sec:experiments}
To demonstrate the efficacy of a GLVM we evaluate it on deformable part models (DPM) by augmenting it with negative parts (GDPM). As discussed in Section \ref{sec:training} adding negative parts changes the optimization framework of DPMs. This includes many noteworthy details about inference, relabeling, data mining, features cache, etc. For details of these modifications refer to Section 4
 in supplementary material. We conducted our experiments of GDPM on two different datasets. Similar to \cite{FelzenszwalbMR08} HOG is used as feature descriptor but the observed trends should be independent of this choice. The results are measured using Average Precision of the PR-curve. We use the PASCAL VOC criterion of 50\% intersection over union for \textit{recall}.\\

\noindent\textbf{PASCAL VOC 2007 Animals. } We first evaluate GDPM on animal subset of VOC 2007. The training set of VOC 2007 contains about $4500$ negative images and a few hundreds ($\sim400$) of positive instances per animal class. The same goes for the test set. We use the same hyper-parameters as used for original DPM \cite{FelzenszwalbMR08}, namely 6 components (3 + 3 mirror components) and 8 parts per component. Our results slightly differ from those reported in~\cite{FelzenszwalbMR08} due to absence of post-processing (bounding box prediction and context re-scoring) and slight differences in implementation. Figure~\ref{tab:voc} summarizes the results for substituting some positive parts in DPM with negative parts. 4 out of the 6 tested classes benefit from replacing positive parts with negative parts. Depending on the class the optimal number of negative parts is different. For PR curves refer to Section 6 
 in supplementary material. \\

\noindent\textbf{Initialization.} Similar to DPM~\cite{FelzenszwalbMR08}, we initialized the location and appearance of positive (negative) parts by finding the sub-patches of the root filter which contained highest positive (negative) weights. However, we observed that negative parts in GDPM are quite sensitive to initialization. This is due to the non-convexity of the optimization. Furthermore,  while the positive label corresponds to a single visual class the negative label encompasses more classes and thus exhibits a multi-modal distribution. But, we are interested in \textit{discrimination} of the positives, thus we only need to model the boundary cases of the negative distribution. Studying these boundary cases for DPM model of \textit{cow} revealed that most of negative samples come from "sheep" and "horse". Initializing GDPM negative parts of cow with a part from similar objects increased the performance from \textbf{21.9} to \textbf{26.7}.\\

\vspace{0.2cm}
\noindent\textbf{Negative Parts.} Although, adding negative parts helps in general, it seems that, unless the parts are carefully initialized, the discovered negative parts with default initialization are not visually meaningful. We analyzed the false positives (FP) of DPM and GDPM detectors similar to \cite{HoiemCD12} and observed that some of the FP reduction in GDPM is from non similar object classes or background. However, when we initialized the negative parts using other classes, the ratio of FP due to similar objects decreased more significantly (See Figure \ref{fig:diagnose}).\\

\input{table1.tex}

\noindent\textbf{Cat Head. } The second task is detection of heads for 12 breeds of cats. The cat head images are taken from Oxford Pets dataset~\cite{Parkhi12a}. For each cat breed the distractor set contains full images (including background clutter) of the other 11 breeds and 25 breeds of dogs. Each cat breed has around 65 positive images and 3500 negative images for training. Test set has the same number of images. This task is particularly hard due to the low number of positive training images and high level of similarity between cats heads which is sometimes hard for humans to distinguish. For sample images refer to Section 5
 in supplementary material. Table \ref{tab:cats} reports the results by comparing DPM and GDPM using various number of positive/negative parts. The aligned average images in \cite{zhu2014averageExplorer} is used as header in Table \ref{tab:cats}. Since the number of positive images in the training set is low, we used one component per detector. It can be seen that GDPM consistently outperform the DPM using different number of parts. For a few classes (\eg Bengal) DPM with 4 positive parts outperform GDPM with 2 positive and 2 negative parts, but as we increase the number of parts to 6, replacing 2 positive parts with 2 negative parts proves beneficial. We observed that adding more parts (more than 6) degrades the results due to over-fitting to the small number of positive images. Negative parts seems to be more robust for the classes where going from total of 4 parts to 6 parts show signs of over-fitting (\eg Birman, Sphynx).\\

%% file: table1.tex
\begin{table}
\renewcommand{\arraystretch}{0.005}%
\setlength{\tabcolsep}{0pt}
\begin{tabu}{cccccccccccccc}
   &\includegraphics[width=28.5pt,height=28.5pt]{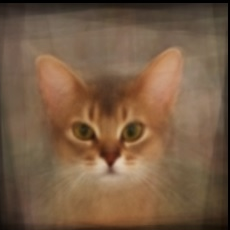}& \includegraphics[width=28.5pt,height=28.5pt]{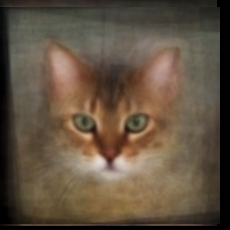} &\includegraphics[width=28.5pt,height=28.5pt]{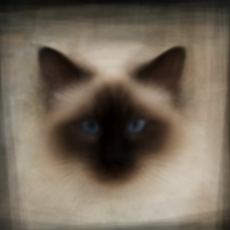}&\includegraphics[width=28.5pt,height=28.5pt]{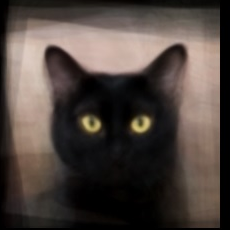}&\includegraphics[width=28.5pt,height=28.5pt]{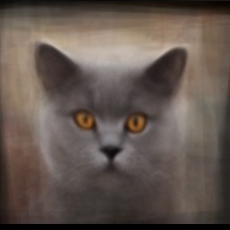}&\includegraphics[width=28.5pt,height=28.5pt]{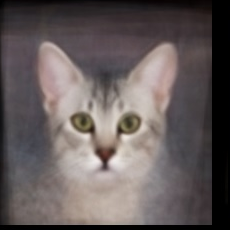}&\includegraphics[width=28.5pt,height=28.5pt]{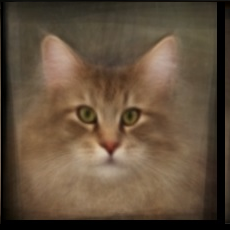}&\includegraphics[width=28.5pt,height=28.5pt]{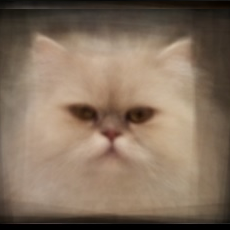}&\includegraphics[width=28.5pt,height=28.5pt]{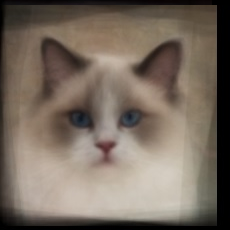}&\includegraphics[width=28.5pt,height=28.5pt]{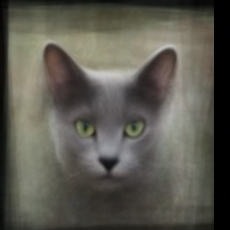}&\includegraphics[width=28.5pt,height=28.5pt]{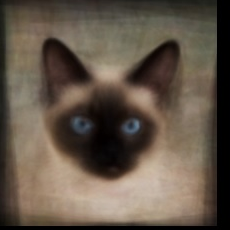}&\includegraphics[width=28.5pt,height=28.5pt]{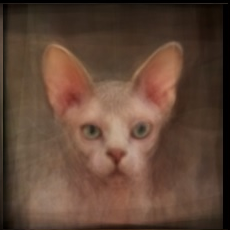}\\[1pt]
   \rowfont{\tiny}
   &\textbf{Abyssinian}&\textbf{Bengal}&\textbf{Birman}&\textbf{Bombay}&\textbf{British}&\textbf{Egyptian}&\textbf{Maine}&\textbf{Persian}&\textbf{Ragdoll}&\textbf{Russian}&\textbf{Siamese}&\textbf{Sphynx}\\
   \rowfont{\tiny}
   &&&&&\textbf{Shorthair}&\textbf{Mau}&\textbf{Coon}&&&\textbf{Blue}&&\\[2pt]
   \taburowcolors 2{white .. lightgray}
   \rowfont{\footnotesize}
   $DPM_{4}$&21.3&12.8&34.5&23.3&32.2&15.8&21.6&28.0&19.4&24.0&29.4&22.7\\[1pt]
   \rowfont{\footnotesize}
  $DPM_{6}$&22.2&12.8&31.4&21.5&31.3&16.5&26.0&29.0&20.6&25.0&30.9&22.0\\[1pt]
  \rowfont{\footnotesize}
   $GDPM_{2}^{2}$&24.3&11.9&\textbf{38.4}&\textbf{23.9}&31.0&19.7&\textbf{27.7}&29.7&\textbf{24.5}&\textbf{29.9}&\textbf{35.9}&\textbf{27.4}\\[0.5pt]
  \rowfont{\footnotesize}
  $GDPM_{4}^{2}$&\textbf{25.5}&\textbf{13.7}&34.0&23.0&\textbf{33.5}&\textbf{20.9}&24.5&\textbf{30.0}&21.9&25.3&30.6&23.2\\
  \hline
\end{tabu}\\
\caption{{\textbf{Cat Head Detection:} \fontsize{\captionfonotsize pt}{0pt}\selectfont  Results comparing DPM and GDPM with different number of positive and negative parts. Sub (super) indices show the number of positive (negative) parts. GDPM consistently outperforms DPM variants. The average cat head images are taken from \cite{zhu2014averageExplorer}.}}
\label{tab:cats}
\vspace{-0.3cm}
\end{table}

%% file: Conclusion.tex
\vspace{-0.3cm}
\section{Conclusion}
\label{sec:conclusion}
\vspace{-0.1cm}
 In this work we proposed a new framework for discriminative latent variable models by \textit{explicitly} modeling \textit{counter evidences} associated with negative latent variables along with the positive evidences and by that emphasized the role of negative data. We further described a general learning framework and showcased how common computer vision latent models can benefit from the new perspective. Finally, we experimented on two tasks of object detection revealing the benefits of negative latent variables. We further observed that expressive power of GLVM crucially depends on a careful initialization. \\
 \noindent We think the proposed GLVM framework yields to fruitful discussions and works regarding various latent variable models.